\documentclass[letterpaper, 10 pt, conference]{ieeeconf}
\IEEEoverridecommandlockouts

\usepackage{amsmath,amssymb,amsfonts}
\usepackage{algorithmic}
\usepackage{cite}
\usepackage{comment}
\usepackage{graphicx}
\usepackage{textcomp}
\usepackage{xcolor}
\def\BibTeX{{\rm B\kern-.05em{\sc i\kern-.025em b}\kern-.08em
    T\kern-.1667em\lower.7ex\hbox{E}\kern-.125emX}}

\newcommand{\note}[2]{\footnote{[\emph{NOTE(#1)}: #2]}}

\begin{document}

\title{Enabling Topological Planning with Monocular Vision
\thanks{*Equal contribution.}
\thanks{All authors are with the Computer Science and Artificial Intelligence Laboratory, Massachusetts Institute of Technology, in Cambridge, USA. \texttt{\{gjstein, cbrad, vpreston, nickroy\}@mit.edu}}
\thanks{This work was supported by the Office of Naval Research and Marc Steinberg under the PERISCOPE MURI Contract \#N00014-17-1-2699. Their support is gratefully acknowledged. V. Preston acknowledges support by a NDSEG Fellowship.}
}

% The \author macro works with any number of authors. There are two
% commands used to separate the names and addresses of multiple
% authors: \And and \AND.
%
% Using \And between authors leaves it to LaTeX to determine where to
% break the lines. Using \AND forces a line break at that point. So,
% if LaTeX puts 3 of 4 authors names on the first line, and the last
% on the second line, try using \AND instead of \And before the third
% author name.

% NOTE: authors will be visible only in the camera-ready (ie, when using the option 'final').
% 	For the initial submission the authors will be anonymized.

\author{
  Gregory J.~Stein$^*$, Christopher Bradley$^*$, Victoria Preston$^*$, and Nicholas Roy
}

\maketitle
%\addtocounter{footnote}{-1}
%===============================================================================

\begin{abstract}
Topological strategies for navigation meaningfully reduce the space of possible actions available to a robot, allowing use of heuristic priors or learning to enable computationally efficient, intelligent planning. The challenges in estimating structure with monocular \textsc{slam} in low texture or highly cluttered environments have precluded its use for topological planning in the past. We propose a robust sparse map representation that can be built with monocular vision and overcomes these shortcomings. Using a learned sensor, we estimate high-level structure of an environment from streaming images by detecting sparse ``vertices'' (e.g., boundaries of walls) and reasoning about the structure between them. We also estimate the known free space in our map, a necessary feature for planning through previously unknown environments. We show that our mapping technique can be used on real data and is sufficient for planning and exploration in simulated \emph{multi-agent search} and \emph{learned subgoal planning} applications.
\end{abstract}

%\begin{IEEEkeywords}
%Topological Planning, Deep Learning, Monocular Navigation, Topometric Maps
%\end{IEEEkeywords}
%%% NOTE (vpreston) : do we include keywords in the formatting?

%===============================================================================

\section{Introduction}
\label{sec:intro}

Autonomous navigation is a ubiquitous problem in the field of mobile robotics.  In order to reduce the number of actions available to an agent, it is often easier to describe the problem, and subsequently generate plans, using the language of topology. In cases where the map is fully known, recent work in the field leverages topological constraints for a variety of different objectives, such as multi-agent search though known maps~\cite{govindarajan2016human}, in which robotic agents are constrained to navigate through different homology classes to search more efficiently. 
%{\color{red}Considering topological properties, like relative homology, has proven helpful when planning through environments that are previously \emph{unknown} to the agent~\cite{kim2013topological}.}

%%%%% For what it's worth, I think I agree with what Chris said about how to change the figure last night -- and as much as the text should stand alone, the figure will still be really compelling at aiding understanding and capturing attention - V
\begin{figure}
    \centering
    \includegraphics{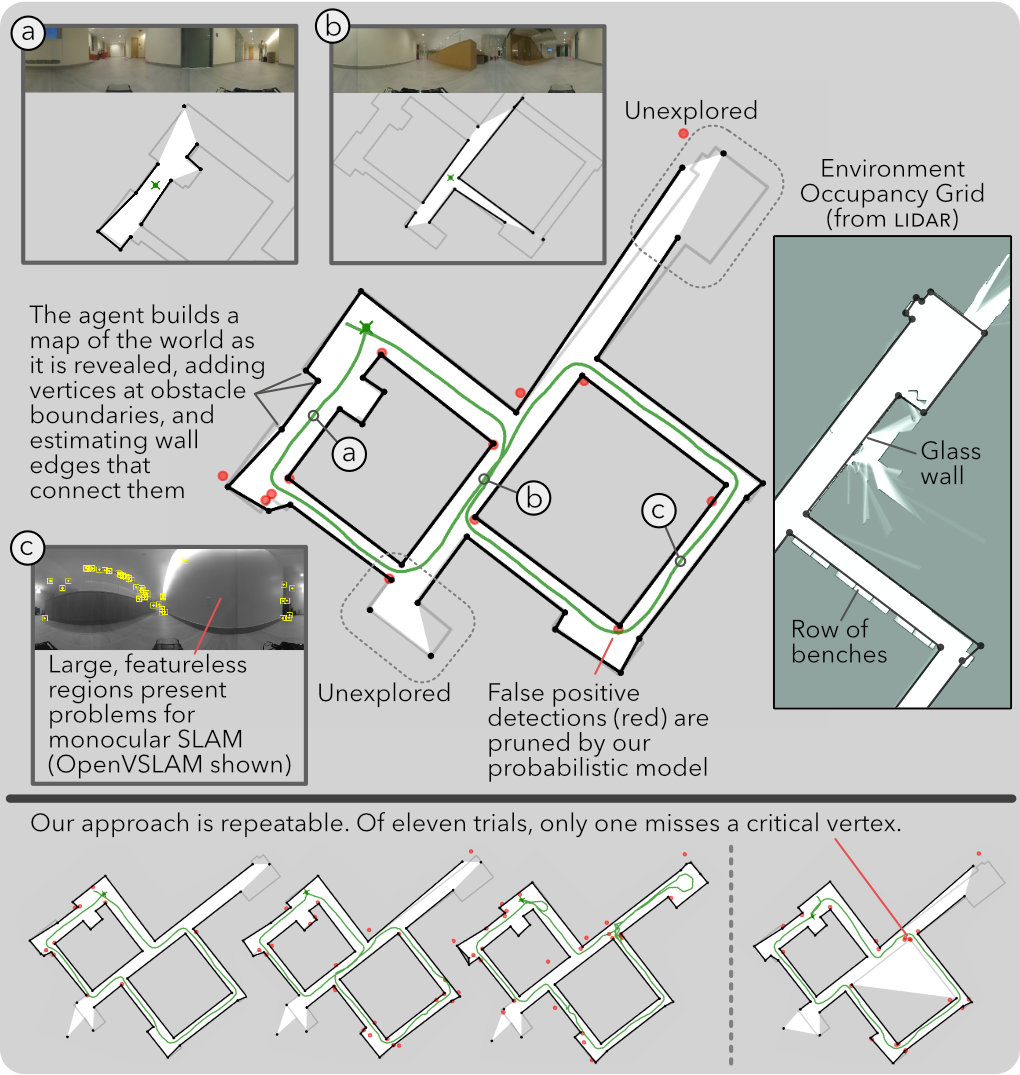}
    \caption{\textbf{Map Building with Real Monocular Camera Data:} Here we show that our procedure builds a sparse map of an environment which is partially observed by a robot equipped with a monocular camera (green line is the robot path). A learned sensor is provided panoramic images (top of panels \texttt{(a)}--\texttt{(c)}) and returns a list of \emph{vertices} which represent corners in the world, in addition to estimates of the \emph{edges} that exist between them. By fusing observations between frames, we can reject detection errors (red circles) and extract the most likely map (solid black lines and circles). Known (shaded white regions) and unknown (shaded gray regions) space is additionally tracked, which can be used by a topological planner. %Our representation avoids adding clutter to the environment (rightmost panel) and was repeatably built over 11 missions with 1 failure case (bottom panel) as a result from poor data association.
    }
    \label{fig:results:real-world-front-page}
\end{figure}

Intelligent decision-making, particularly in the context of navigating unknown space, can be modelled as a Partially Observable Markov Decision Process (\textsc{pomdp})~\cite{Kaelbling:1998:PAP:1643275.1643301}.  Topology provides a means of specifying \emph{high-level} actions available to the agent (e.g., ``go left around the building'' or ``enter unknown space through this boundary''), allowing the planner to reduce the computational challenges of the optimization problem implicit in the \textsc{pomdp}. Planning under topological constraints reduces the space of actions, making it easier to plan~\cite{kim2013topological,park2015homotopy} or encode heuristic or learned priors~\cite{Korb2018,stein2018learning}. Critically however, to plan with topological constraints, a robot must be able to extract them from its environment.

% "frontier-based exploration would not be suitable with our monocular SLAM system" - QuotED

%Unfortunately, the maps produced by monocular \textsc{slam} systems are most frequently designed and tuned for accurate localization\note{vpreston}{...rather than mapping...}, resulting in point clouds of sparse visual features whose noise limits its feasibility for the construction of maps that support planning under topological constraints. Furthermore, such point clouds cannot easily distinguish between known and unknown space, an essential component for navigation through previously unknown environments.

The ubiquity of cameras and the richness of information vision provides makes monocular images an attractive candidate for both informing high-level autonomy and building the maps autonomous agents use to represent their environment. Monocular, feature-based methods have proven to be effective tools for localization and navigation~\cite{cadena2016past} and have shown promise in building occupancy maps sufficient for exploration of small environments~\cite{von2017monocular,mostegel2014active}. These approaches build sparse point-clouds of visual features, and bin them into voxels to represent obstacles which a robot could use for path planning. However, such approaches have limitations when planning with topological constraints in general; spurious detections can block potential paths, and a lack of detections in featureless regions (e.g., a white wall in a hallway) can lead a robot to believe free space exists where it does not. Resolving the discrepancies between the point cloud and the true underlying geometry of an environment is a challenge for such systems. Furthermore, due to the high dimensionality of point-cloud-based \textsc{slam}, robustly eliminating erroneous features is an open problem~\cite{cadena2016past}. 

Planning with topological constraints necessitates a representation which is robust to the types of failures described: all obstacles must be added so that dead ends can be detected and planned around, while spurious obstacles (that may block paths or add unnecessary options) should be eliminated. Specifically for the exploration problem, even state-of-the-art work via monocular vision acknowledges the challenges (and failures) of extracting frontiers from their dense map reliable enough to enumerate the different ways the agent can enter unknown space during planning~\cite{von2017monocular}. Relatedly, most planning strategies that leverage \emph{homology} or \emph{frontiers} rely on ideal sensors~\cite{tovar2003optimal,Korb2018} or specialized hardware, like a laser range finder, to build a map and generate plans~\cite{yamauchi1997frontier,tian2018rescue}. 

During exploration and navigation, planning with a map may be done via computation of a \emph{visibility graph}~\cite{lozano1979algorithm}, which is a minimal, optimal representation for planning in a known environment and can be extracted directly from a map. Maps generated using sparse features contain \emph{clutter} or voids of features (e.g., in low-texture regions), which yield visibility graphs that are not useful for planning. Recent progress in deep learning for computer vision affords opportunities to overcome these faults comparatively easily by only adding sparse features where necessary for developing useful graphs for planning, without reliance on surface textures and by ignoring clutter or outlying detections.

The contribution of this paper is to extend recent work on planning with topological constraints to maps constructed with monocular vision, with an emphasis on planning through unknown environments. In this work, we present a novel map representation built from sparse detections from monocular vision, which is robust to noise in a way that enables navigation and exploration using topological constraints in both simulated and real-world environments. From our representation, we compute a visibility graph corresponding to the sparse, clutter-free structure of the known environment from which we can efficiently navigate using topological constraints. To build our map, we use a Convolutional Neural Network (\textsc{cnn}) to detect the edges of locally visible structure, which are fused to estimate regions of known and unknown space, and track obstacles that define topological constraints. 

% To demonstrate the utility of our representation for navigation and exploration, we focus on two applications

We first study \emph{multi-agent search}; our method succeeds in 99\% of trials across two simulated environments, demonstrating the robustness of our procedure (Sec.~\ref{sec:search_and_rescue}). We additionally show that our representation enables \emph{visual learned subgoal planning}, demonstrating that our map is sufficiently stable to enable learning-based selection of topological actions for improved goal-directed navigation (Sec.~\ref{sec:results:lsp}). Finally, we show that our work extends beyond simulation by building maps in three real-world environments (Sec.~\ref{sec:robot_experiments}).

\section{Planning with Topological Constraints}
\label{sec:planning-theory}

In this section, we examine requirements of topological planners and motivate our map representation. In order to efficiently navigate through both known and unknown environments, topological constraints can be employed to reduce the space of actions available to a robot. Two trajectories are said to be in different \emph{homology classes} if the area they bound contains obstacles; planning approaches may use homology---a topological property---to constrain possible actions the agent can take (e.g., ``go left around the building'').
% Planning approaches that use \emph{homology classes} 
% Planning approaches that use \emph{homology classes} compute the net angle a given trajectory accumulates around each obstacle during travel, which constrains possible actions the agent can take (e.g., ``go left around the building''). 
Storing a single point per contiguous obstacle is a minimally sufficient representation for computing the homology class of a trajectory~\cite{bhattacharya2012topological}. Of interest in this paper is planning through unknown environments during map construction, for which \emph{relative homology}~\cite{pokorny2016topological} considers only partially revealed obstacles. \emph{Frontiers} are boundaries between free and unknown space, and constrains planning to relative homology class. \emph{Frontier-based planning} restricts trajectories to pass through the frontier of interest, thus trajectories are guaranteed to have unique relative homology. To use planners based on relative homology (or map frontiers), we need a representation that tracks both obstacles and free space.
% Storing a single point per obstacle on the boundary is minimally sufficient for computing the relative homology of a trajectory leaving known space.

The \emph{Gap Navigation Tree}~\cite{tovar2007gapnavigation,tovar2004gapnavigation} is a mapping technique for building a sparse, minimal, tree-structured graph of the world by detecting \emph{gaps}, discontinuities in depth, and maintaining a list of unexplored branches for later exploration. \emph{Gaps} store the high-level actions available to the robot, yet their practical utility is limited by sensor noise~\cite{murphy2008gntlaser} and, due to its tree structure, Gap Navigation is limited to simply-connected environments. Moreover, gap locations are viewpoint-dependent, creating challenges for data association and robust map construction. To overcome these challenges, we approximate the environment as a polygon, so that the location of gaps become viewpoint independent, enabling reliable data association. As such, our sensor detects polygonal \emph{vertices} in view of the agent; by localizing vertices and estimating the presence of walls or other impassible obstacles that connect them, our agent is capable of reconstructing the polygonal representation of its environment as it travels (see Sec.~\ref{sec:mapping-theory}). 
% To overcome these challenges, we need to identify markers that are position independent. 
% By approximating the environment as a polygon, the location of gaps become viewpoint independent, thereby enabling data association.
%To overcome these limitations, we instead approximate obstacles as polygons and detect their vertices. By localizing vertices and estimating the presence of walls or other impassible obstacles that connect them, our agent is capable of building a polygonal representation of its environment as it travels (see Sec.~\ref{sec:mapping-theory}).

Our map representation consists of the vertices and edges that define obstacles and an estimate of the free space observed by the robot. Planning within this representation is straightforward via computation of a visibility graph~\cite{lozano1979algorithm}, which is minimal and optimal in a known polygonal environment. For exploration tasks, the representation of free space can be used for frontier-based planning by extracting frontiers from the difference between map edges and the boundary of free space. Additionally, we can easily compute the relative homology class of a trajectory from our representation by tracking partially revealed obstacles, making it sufficient for planning with topological constraints.

\begin{figure}[t]
    \centering
    \includegraphics{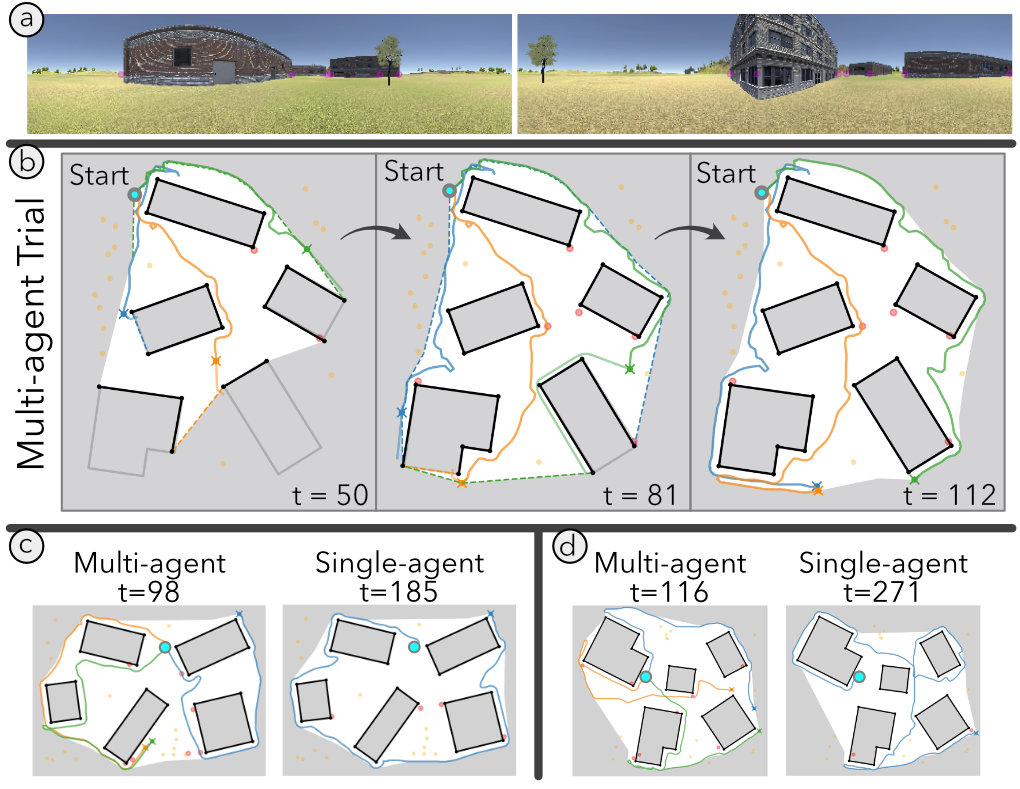}
    \caption{\textbf{Search in Simulated Office Park:} Here we show three representative search scenarios in a simulated outdoor environment \texttt{(a)} populated with buildings and trees. In \texttt{(b)}--\texttt{(d)} the progress of multiple agents is shown, where the agents are constrained to pursue unique homology classes. We successfully identify and map all buildings in 99 of 100 simulated trials, illustrating the robustness of our technique. See Sec.~\ref{sec:search_and_rescue} for more details.}
    \label{fig:results:sar:outdoor}
\end{figure}

\begin{comment}

Things that need to get done:
- massively simplify the discussion of what is currently called 'vertex orientation'
- Collapse the 'intro', since much of the notation and such belongs in an earlier section
- Make clearer why we detect vertices, as opposed to directly reasoning about walls.
- move the CNN description to the end of the section (and just refer to it earlier on). "Details about the CNN can be found in Sec. ??"
- 

Outline:
- [0] Define the noisy map: vertices, walls between vertices, known space definition; sensor is learned from vision(?)
- [1] Posterior over maps
- [2] Data Association (? should this be in the posterior over maps section) 
- [3] Sampling (probably unchanged)

Misc.
- 

\end{comment}

\section{Probabilistic Map-building from Learned Monocular Sensing}
\label{sec:mapping-theory}
Our agent is equipped with a learned sensor (Sec.~\ref{sec:theory:learned-sensor}) that returns a list of detected vertices in view of the agent and, for each detection, the likelihood an obstacle exists between it and each of its neighbors (e.g., if the vertex and its neighbor are endpoints of a shared wall). We learn to identify where the robot can and cannot travel in order to explore its environment, and do this by detecting the boundaries of untraversable obstacles.
%A description of the Convolutional Neural Network (\textsc{cnn}) used to detect vertices from panoramic images is included in Sec.~\ref{sec:theory:learned-sensor}. 
Our map representation therefore stores three types of information:
(a) \textbf{vertices}, each a cluster of vertex detections; 
(b) \textbf{edges}, connections between two vertices which form an untraversable boundary; and a
(c) \textbf{known space polygon}, a polygon defining what space has been observed to be free.

% Data association for vertices
\paragraph{Vertices} Each vertex detection in an observation $z^{(i)}$ is compared to the vertices that exist in the map in order to determine how each detection should be associated. Either a detection is matched with an existing vertex, or it is added as a new vertex to the map with 2D pose and covariance. The Mahalanobis distance~\cite{murphy2008gntlaser} is used to associate vertex detections between observations in order to cluster detections into vertices. We additionally apply a mutual exclusion constraint~\cite{1241885_thrun_data} such that no two vertex detections from a single observation are associated with the same vertex. Successful associations update the position and covariance of vertices using a Kalman filter. %If a vertex is unmatched, it is added to the map, represented by a position in 2D, and a covariance that indicates the uncertainty about that pose. %For each vertex detection that was associated with an existing one, the position and covariance are updated via a Kalman Filter.

%Finally, when a detection can be associated with multiple vertices, association priority is given to map vertices with low probability of being a false positive; the impact of this heuristic is generally small, but empirically reduces the influence of false detections during map building.

%\paragraph{Polygonal Vertices} Mahalanobis distance~\cite{murphy2008using} is used to associate vertex detections between observations in order to extract map vertices. A new vertex detection is \emph{associated} with a polygonal vertex when the polygonal vertex lies within one standard deviation of the detection's covariance. We additionally apply a mutual exclusion constraint~\cite{1241885_thrun_data} such that no two vertex detections from a single observation are associated with the same map vertex. Successful associations update the position and covariance of map vertices using a Kalman filter. If no matching map vertex exists, a new vertex is added to the map. Finally, when a detection can be associated with multiple vertices, association priority is given to map vertices with low probability of being a false positive; the impact of this heuristic is generally small, but empirically reduces the influence of false detections during map building.

% Edges (directly detection is hard)

\paragraph{Edges} Edges between vertices impose topological constraints necessary for planning. However, directly detecting the structural connectivity of an environment using vision is a challenging problem~\cite{greene2018flame}. For each vertex detection our sensor instead returns a likelihood that an untraversable obstacle exists between each of its covisible neighbors (an ``edge likelihood''). For an observation $z^{(i)}$, we average these likelihoods for each pair of detections to define the probability an edge exists between the vertices; over many observations we average these probabilities to determine whether or not an edge exists in the map.
% We then average these probabilities for each pair of detections in an observation to define the probability an edge exists between vertices, which is similarly averaged over a sequence of observations.

%about whether it is connected via an obstacle to each of its covisible neighbors. The average of these defines the edges in our map.

% "Perfect" known space
\paragraph{Known Space Polygon} Each sensor observation $z^{(i)}$ reveals a \emph{star-shaped} region surrounding the agent $s(z^{(i)})$. The function $s$ constructs a polygonal region by sorting the vertex detections by the angle at which they were observed. The union of these polygons, $S = \bigcup^{N_\text{obs}}_{i=0} s(z^{(i)})$ for all sensor observations $N_{obs}$ defines the space revealed by the agent.

When the sensor is perfect, the map representation can be built directly from the descriptions provided. In practice however, the sensor is noisy---both vertex and edge detections will not only be imperfect, but false positive and negative detections may also be introduced.
% [Discuss how the map is built].
% The vertices and walls are estimated from the sparse vertex detections and fused using [data association]. Yet the detector is fundamentally noisy, and some of the estimated vertices may be false-positive detections. As such, 
In order to handle sensor noise, we define a probabilistic model from which we can assess the likelihood of a proposed map; the posterior likelihood of maps is discussed in Sec.~\ref{sec:theory:posterior}. To propose and evaluate the posterior over maps, we sample over the set of polygonal vertices and their most likely edges, then compute free space in the presence of noise. This discussion is included in Sec.~\ref{sec:theory:sampling-possible-maps}.

\subsection{Vertex Detection with Convolutional Neural Networks}
\label{sec:theory:learned-sensor}

In order to detect vertices, we train a \textsc{cnn} to estimate the vertex and two edge likelihood terms. The network takes a $128\!\times\!512$ pixel RGB panoramic image and returns the 3 likelihoods over a $32\!\times\!128$ grid, where each point corresponds to a range-bearing coordinate from the sensor\footnote{To recover a discrete set of vertices from the grid based output produced by the \textsc{cnn}, we use a peak-detection procedure with a threshold of 0.5.}. We use a fully-convolutional encoder-decoder network structure composed of blocks. Each encoder block consists of 2 convolution layers with $3\!\times\!3$ kernels, followed by a batch norm operation and a ReLU activation function, and terminated by a $2\!\times\!2$ max-pooling operation. Decoder blocks are similar, however the final convolutional layer has stride 2 (so that the output is upsampled) and the max-pooling operation is eliminated. The network consists of 5 encoder blocks, with output channels $[64, 64, 128, 128, 256]$, and 3 decoder blocks, with output channels $[128, 64, 64]$. The output of the final decoder layer is passed through a final $1\!\times\!1$ convolutional layer with 3 output channels, corresponding to the 3 likelihoods. A weighted cross-entropy with an empirically chosen positive-weight 8 is used as the loss function for the vertex likelihood, so that the sparse positive detections are not overwhelmed by the negative background. The edge likelihoods each have a sigmoid cross-entropy loss, yet are masked by the training vertex label so that loss is only non-zero where a vertex exists. The edge likelihood loss is weighted by a factor of 1/16 versus the vertex likelihood so that the network prioritizes whether a vertex exists before trying to estimate its properties. The Adam optimizer is used to train the network for 100k steps, with an initial learning rate of 2.5 and a learning rate decay of 0.5 every 10k steps.

\subsection{Generating Proposal Maps}
\label{sec:theory:sampling-possible-maps}

%Each detected vertex is either associated with an existing vertex, or forms a new one
Associating detected vertices across many observations results in a set of \emph{potential vertices}, of which only a subset may appear in the final map. By iteratively including and removing potential vertices, and the walls connected to them, we explore the space of possible maps via sampling\footnote{We randomly sample over vertices uniformly.}.
% To explore the space of maps, we treat the inclusion of each vertex and edge as binary variables and sample this space to compute the maximum likelihood map.
%The output of each observation is a set of \emph{vertex detections} which, will be associated with one of the vertices that exist in the map. Due to noise in the detector, some of these vertices may be false-positive detections. Similarly, the detector may have failed to recognize a vertex, marking a false-negative detection. Perhaps most significantly, errors in the estimated position of a \emph{vertex detection} might cause it to be associated with the wrong vertex in the map. These compounding sources of noise may lead to a map where there exist vertices or walls which do not belong.
% \todo{nick}{This sentence does not logically follow. The last page was about verticies and learned subgoal properties [gjstein: no longer true]. You need text to explain the maximum a posteriori map and inference procedure (?) that requires differentiability.}\todo{nick}{This pragraph needs rewriting for logical flow. The sentences need to follow logically from one another.}Since the space of possible maps is non-differentiable, we must sample over map space to compute the maximum likelihood map. The output of data association is a set of \emph{potential vertices}, of which only a subset may appear in the final map. By iteratively including and removing potential vertices, and the walls connected to them, we explore the space of possible maps. 
Computing the map also requires computing the known space polygon and, as we change the vertices and edges included in the map, the known space must be updated. To construct this polygon for a given set of vertices and edges, we first compute a hypothetical observation $z_h$ by ray casting against the proposed structure ($z_h$ is what the robot would see if the proposed structure were accurate). However, this detection does not include unobserved obstacles. As such, we instead compute a \emph{conservative estimate} of known space $S_c$ when proposing maps: the intersection of the polygons computed from both the real and hypothetical sensor observation:
\begin{equation}%\textstyle
    S_c = \bigcup^{N_\text{obs}}_{i=0} S_c^{(i)} = \bigcup^{N_\text{obs}}_{i=0} \left( s(z^{(i)}) \cap s(z^{(i)}_h) \right)
\end{equation}
Without this conservative approach, the agent may incorrectly mark unobserved space as explored, leading to incomplete exploration.

We probabilistically accept proposed maps using an \textsc{mcmc} criteria and the map posterior. After a fixed number of samples (50 samples were used for all experiments presented in this work) we return the maximum likelihood map. In general, sampling requires iterating over the inclusion of walls in the proposed map. In practice, we use a heuristic to determine whether a wall exists: we compute an approximate marginal edge likelihood by computing individual observations of the edge likelihood using co-visible vertices in the \emph{real} detections (rather than the \emph{conservative} detections). We have found this heuristic sufficient for effective map-building and planning and considerably reduces the search space.

\subsection{Defining the Posterior Over Maps}
\label{sec:theory:posterior}
Given $N$ observations $\{z^{(i)}\}_{1 \leq i \leq N}$, the posterior distribution over maps $m$ can be expanded using Bayes Rule:
\begin{multline}\textstyle
\log P(m | z^{(1)}, \cdots, z^{(N)} ) = \\\textstyle \sum_{i=0}^N \log P(z^{(i)} | m) + \log P(m) - \log P( z^{(1)}, \cdots, z^{(N)} )
\label{eq:posterior}
\end{multline}
where we have assumed that the sensor observations are i.i.d. to factor $\log P(z^{(1)}, \cdots, z^{(N)} | m)$. The final term in Eq.~\eqref{eq:posterior}, the likelihood of a sensor measurement, is a normalizer and can be ignored. The prior distribution over maps, $P(m)$, is intractably difficult to obtain in practice. However, we can instead incorporate heuristic priors about the map, namely biases against the addition of new vertices and walls, which mitigates the impact of false positive detections.

% OLD
% By using a conservative estimate of known space, we can identify false positive and negative detections: any corners that appear in the real observation that are not in the hypothetical observation are considered false positive detections, and corners that appear in the hypothetical observation but were not in the real observation are considered false negative detections. 

To evaluate the likelihood of each observation given a proposed map, we compare the sensor observation against the conservative space estimate for individual observations $S_c^{(i)}$. When evaluating the likelihood of a sensor observation $P(z^{(i)} | m)$, we reason about the likelihood of the vertices $P_v(z^{(i)} | m)$ and edges $P_e(z^{(i)} | m)$ independently. The contribution from the vertices depends only on the number of false positive and false negative detections: any real vertex detections that do not appear in $S_c^{(i)}$ are considered false positive detections and any vertices within $S_c^{(i)}$ that were not detected by the sensor are considered false negatives. Therefore, $P_v(z^{(i)} | m) = R_\textsc{fp}^{N_\textsc{fp}} R_\textsc{fn}^{N_\textsc{fn}}$, where $R_\textsc{fp}, R_\textsc{fn}$ are the rates of false positive/negative detections and $N_\textsc{fp}, N_\textsc{fn}$ are the number of observed false positive/negative detections. 

% Edge likelihood
The likelihood contribution from the edges $P_e(z^{(i)} | m)$ also makes use of the conservative space estimate $S_c^{(i)}$. For each vertex our sensor returns the likelihood it is connected via an obstacle to each of its covisible neighbors. The average of these defines the edges in our map. False negative detections, which do not have an associated real vertex detection, are assigned a uniform prior over edge likelihood.

Our posterior codifies a number of behaviors we would expect to see during our map-building process. First, the posterior allows us to omit false positive vertices in a principled manner. Since false positive vertices can result in unwanted false negative observations as the robot travels, it ultimately makes it more likely that the vertex does not appear in the underlying geometry. Second, the existence of a wall depends on both the detected edge likelihood \emph{and} its ability to occlude vertices. By occluding portions of the map, the edges also influence the number of false positive and false negative detections recorded during the evaluation of map likelihood.

\begin{figure}[t]
    \centering
    \includegraphics{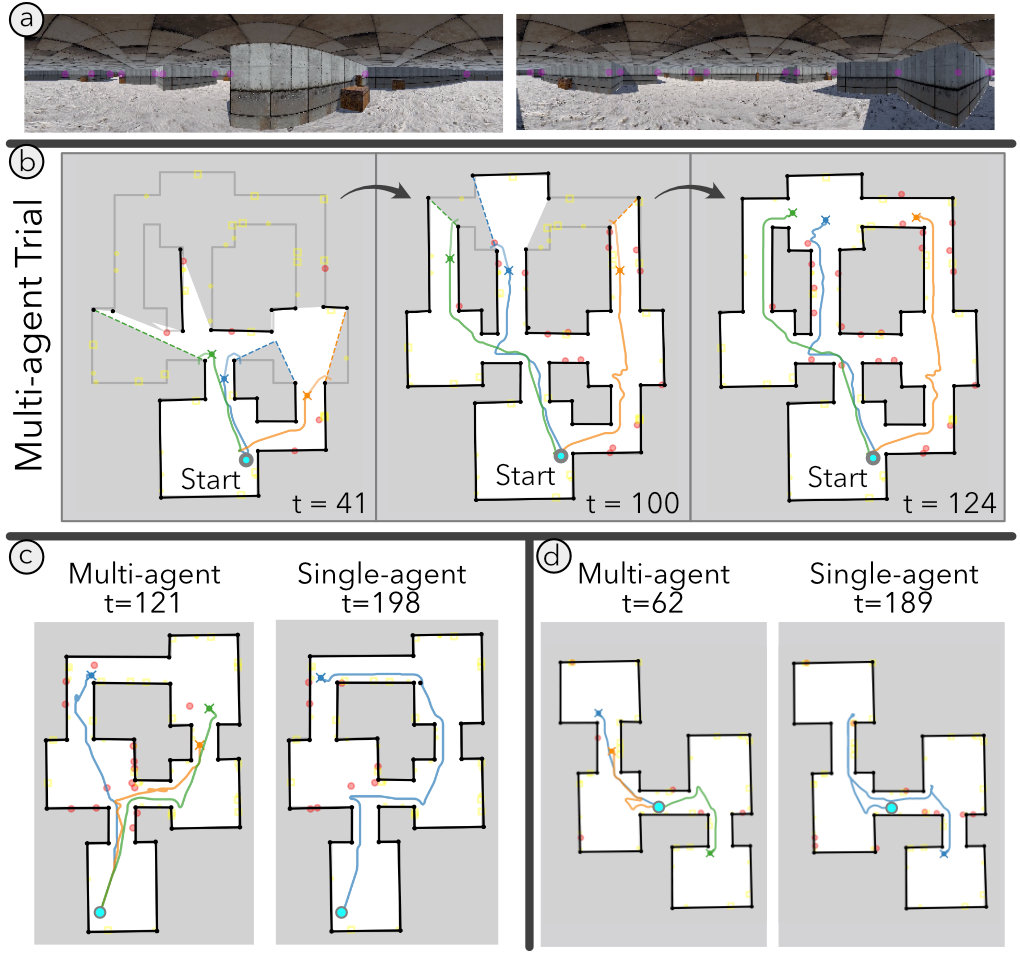}
    \caption{\textbf{Search in Simulated Indoor Environment:} Here we show three representative search scenarios in a simulated indoor environment populated with random clutter \texttt{(a)}. In \texttt{(b)} the progress of multiple agents is shown, where the agents are constrained to enter unknown space via a unique frontier. In panels \texttt{(b)-(d)}, just as in the \emph{Office Park} trials in Fig.~\ref{fig:results:sar:outdoor}, multi-agent teams complete exploration in roughly half the time (on average) that single agents use.}
    \label{fig:results:sar:catacombs}
\end{figure}

\section{Multi-Agent Search}
\label{sec:search_and_rescue}
% \todo{gjstein}{motivate search and rescue; cite a couple papers; define metric(s) for success: reliable coverage and completion time.}

% We would like to make this section a "utility" section -- our representation can enable traditional methods of topology planning, and it does it well. Demonstrated by:
% - multi/single agent statistical results
% - qualitative behaviors (natural divide-and-conquer behaviors for multi agents)
% This is impressive because
% - other methods can't or have a hard time extracting useful frontiers (cite lit)
% - learning depth well with a monocular camera is challenging
% - robust to noise ("laser-like maps")
% - able to ignore clutter ("clean maps")

To demonstrate the reliability of our approach, we test our map-building procedure with \emph{multi-agent search}, in which agents must be able to maintain an accurate estimate of observed space in order to reach an unseen goal. Multi-agent search strongly benefits from planning with topological constraints, so that each agent is encouraged to explore different regions of space and reveal the environment more quickly~\cite{govindarajan2016human}. The challenges in detecting features in textureless regions and accurately resolving depth in cluttered regions make traditional monocular \textsc{slam} approaches impractical for this task~\cite{von2017monocular,mostegel2014active}. 

We conduct 200 simulated experiments across two environments we created in the Unity Game Engine~\cite{unity}: (1) an \emph{Indoor Environment}, a small labyrinth of rooms connected by corridors, and (2) an \emph{Office Park}, an outdoor environment of randomly placed buildings and trees. Training data is generated via simulated travel through a subset of environments not included in the test set.
% For single-agent planning, no topological constraints are enforced and the agent simply seeks the closest region of unknown space.
To plan with multiple agents, a joint planner encourages the agents to minimize the net cost of entering unknown space yet via different frontiers, thereby ensuring that agents select trajectories in different relative homology classes. %; this demonstrates how reliably we can compute high-level topological actions with our map.
Map construction with multiple agents is straightforward: observations are received in sequence from each of the three agents and added to the map in turn. Since the probabilistic model treats each measurement as i.i.d., no modifications are necessary to our map-building procedure. 

\begin{table}[t]
    \centering
    \caption{Performance metrics (averaged) for the search task}
    \begin{tabular}{r|ccc|ccc}
    & \multicolumn{3}{c|}{\textbf{Indoor Environment}} & \multicolumn{3}{c}{\textbf{Office Park}}\\
    & \!Success\! & Steps & IoU &
    \!Success\! & Steps & IoU \\\hline
        Multi-agent & 99/100 & \textbf{137.6} & 0.979  &  100/100 & \textbf{128.0} & 0.990 \\
        Single-agent & 99/100 & 230.6 & 0.979  &  97/100 & 254.1 & 0.990
    \end{tabular}
    \label{tab:sar:results_table}
\end{table}

% In the latter, multiple agents work in tandem to search the environment and construct a joint map. The process for constructing the map with multiple agents is straightforward: observations are received in sequence from each of the three agents are added to the map in turn. Data association still relies on Mahlanobis distance to add observations and, since the probabilistic model treats each measurement as i.i.d no modifications are necessary. 

% To demonstrate that our mapping procedure is sufficiently robust to support planning with topological constraints, we implement \emph{single-} and \emph{multi-agent search and rescue}. 

We report performance in terms of coverage---the agent's map of the world should match the underlying map. As such, we use \emph{Intersection over Union} (IoU) between the final reconstructed map and the true map as a measure of coverage\footnote{We compute IoU differently in the outdoor environment, since we should not expect that the agent should mark all space around the buildings as free. Instead, we compute IoU with respect to the placement of the building obstacles. Mapping is perfect when all buildings are detected and in the correct locations and that no spurious buildings are added.}. Examples of our results are shown in Figs.~\ref{fig:results:sar:outdoor} and~\ref{fig:results:sar:catacombs}, and a data table summarizing the results of all simulated trials---100 randomized maps for each simulated environment---are in Table~\ref{tab:sar:results_table}. Of our 200 trials, 198 succeed in completely exploring their environment, achieving an IoU above 0.95. The two remaining trials miss a single wall and the agents became stuck in place. This results in identical input images fed to our model, which will produce identical outputs. This violates the i.i.d. assumption of our sensor model and the maximum likelihood map may not match the underlying environment. This does not occur often and, despite this occasional limitation, our representation is sufficient for mapping and planning in unknown simulated environments with monocular vision, even in the presence of sensor noise. Data from the 200 multi-agent trials can be found in Table~\ref{tab:sar:results_table}; we show performance alongside an additional 200 trials from single-agent planning, reinforcing that imposing topological constraints with our map enables more efficient search of unknown environments.

\section{Visual Learned Subgoal Planning}
\label{sec:results:lsp}

\begin{figure}
    \centering
    \includegraphics{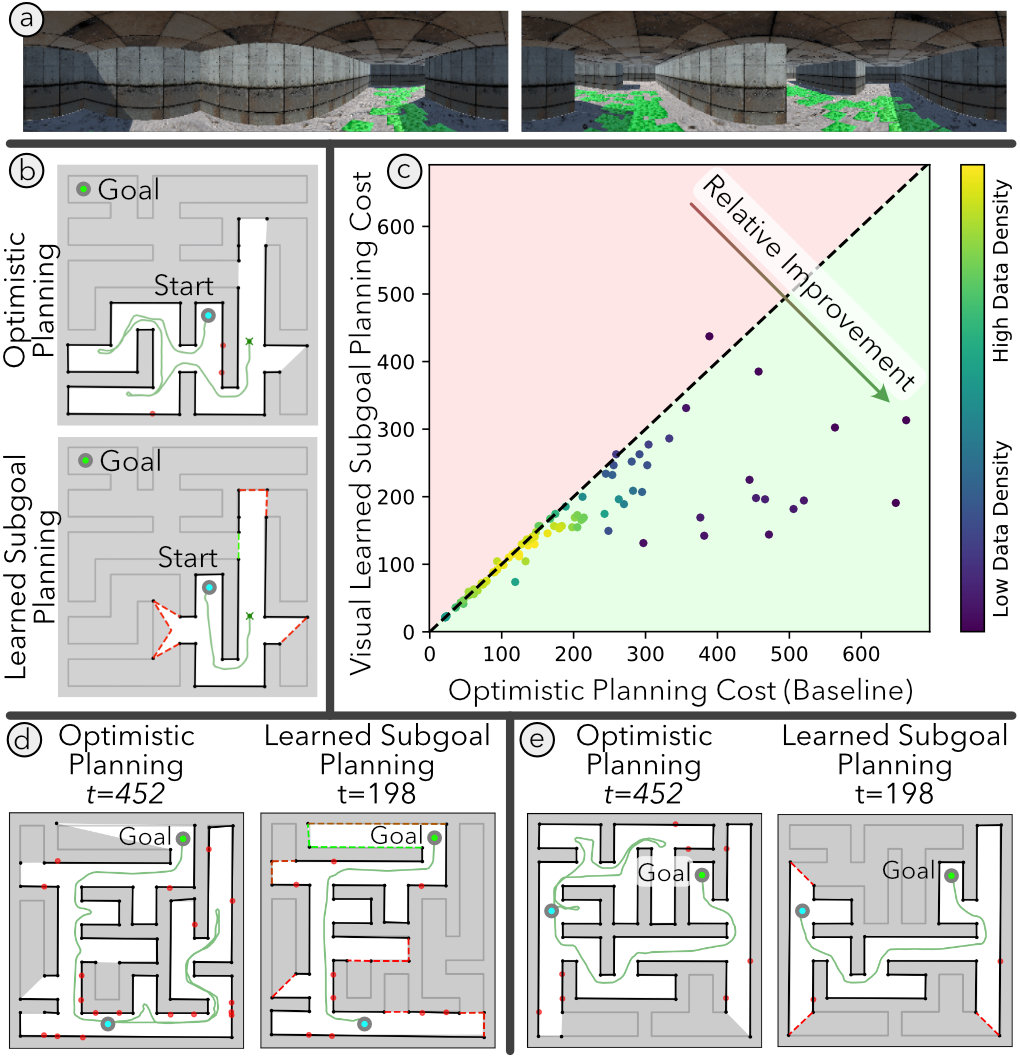}
    \caption{\textbf{Optimistic and Learned Subgoal Planning in Simulated Guided Mazes:} Our map representation is used by an optimistic planner and a learned subgoal planner (\textsc{lsp}) that are compared in 200 simulated guided maze environments in which green ``breadcrumbs'' lead to the goal \texttt{(a)}. In panel \texttt{(b)}, snapshots of the two planners in progress are shown, wherein the optimistic planner is greedily searching nearby paths and the learned subgoal planner is pursuing frontiers marked with dashed lines (red are unlikely to reach the goal, green are more likely to reach the goal). Panels \texttt{(d, e)} show completed missions for each planner. As shown in panel \texttt{(c)}, robots using the learned subgoal planner travel a total of 25.6\% less distance than those using the optimistic planner. This result highlights that our method is not only suitable for navigation, it is stable enough to support \textsc{lsp}.}
    % \caption{A comparison between the cost-to-go of the optimistic planner baseline and our subgoal planner for 200 simulated trials in the Guided Maze Environment. The blue line represents the calculated fit line of data points, each corresponding to a simulated trial. The Learned Subgoal Planning agent travels a total of 22.3\% less distance than the optimistic baseline over these trials, which is shown in the scatter plot by the fit line being in the green colored region. In plots of the subgoal planner's path, boundaries associated with each subgoal are colored from green to red so as to visualize the estimated likelihood that it will lead to the goal.}
    \label{fig:results:vlsp_maze_results}
\end{figure}

Our final simulated experiments demonstrate that our representation enables Learned Subgoal Planning (\textsc{lsp})~\cite{stein2018learning} without the need for specialized hardware. \textsc{lsp} involves estimating properties of contiguous boundaries between free and unknown space (frontiers), including the likelihood the goal can be reached through the boundary of interest, and uses these properties to approximate the expected cost of topological actions. In~\cite{stein2018learning}, the authors showed that their approach is a computationally tractable means of computing expected cost and demonstrated improvements for planning through different types of environment. Because the learning procedure relies upon stable, well-defined boundaries between free and unknown space, previous work on \textsc{lsp} relied on a planar laser scanner to build an occupancy grid map. We demonstrate here that our map representation from monocular vision is sufficiently reliable to enable \textsc{lsp}, thus showing that vision can be used for both enumerating the high-level actions available to the agent during exploration and deciding between these actions.
% which requires stable, well-defined frontiers in order to operate effectively. 

The \textsc{lsp} experiments require three frontier properties~\cite{stein2018learning}---the likelihood a frontier will lead to the goal, the expected cost of reaching the goal (if the goal can be reached, and the expected cost of exploration (if the goal cannot be reached)---that we estimate alongside the vertex and edge likelihoods by adding output channels to our network architecture described in Sec.~\ref{sec:theory:learned-sensor}. A few other changes to the neural network are made when estimating frontier properties. The frontiers can overlap in range-bearing space, so we add an additional decoder block to create a higher-resolution output. To ease initialization of this larger network, we found it more effective to use leaky-ReLU for the convolution layer activations. The goal location, needed to estimate frontier properties, is encoded into a 2-channel image and appended to the input of the third encoder block; each pixel corresponds to $[r_g \sin(\theta_g), r_g \cos(\theta_g)]$, where $r_g$ is the relative range and $\theta_g$ is the relative angle of the goal versus the bearing of the pixel of interest. See \cite{stein2018learning} for a complete discussion of the frontier properties.

To evaluate performance, we created a visually-oriented variant of the \emph{guided maze} environment from \cite{stein2018learning}. Each randomly-generated map is a minimum spanning tree maze, yet the path between the start and goal is marked by a green path on the ground. An agent using the learned subgoal planning algorithm should learn that the highlighted path indicates actions that are more likely to lead to the goal and preferentially pursue those. Vision therefore provides a strong signal for the correct path. 

We conducted 400 simulated trials in our \emph{guided maze} environments: 200 in which the agent builds the map and uses the \textsc{lsp} algorithm to select the lowest expected cost action, and 200 using a naive baseline that plans as if all unknown space is free. All 400 trials reach the unseen goal, speaking to the reliability of our map-building and navigation procedure. Results showing a side-by-side comparison of the \textsc{lsp} procedure versus the naive baseline are shown in Fig.~\ref{fig:results:vlsp_maze_results}. As expected, the learned planner matches or outperforms the baseline planner in nearly all trials (25.6\% reduction in net cost), highlighting that our representation is not only able to support navigation, but also is stable enough to enable learning the properties necessary for \textsc{lsp}.

\begin{figure}[t]
    \centering
    \includegraphics{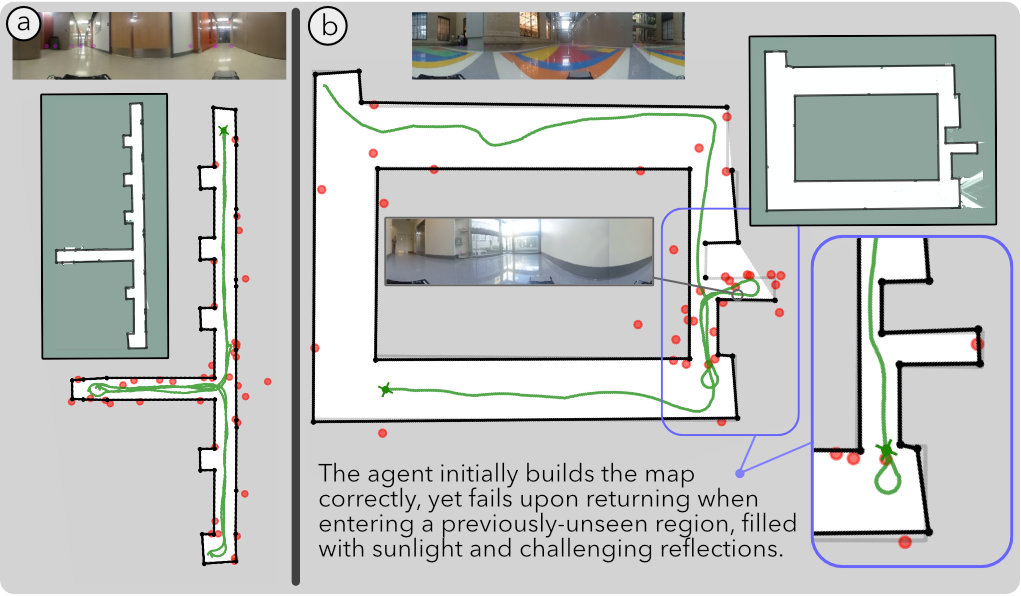}
    \caption{\textbf{Additional Real-World Mapping Trials:} Beyond the real-world result shown in Fig.~\ref{fig:results:real-world-front-page}, we show here mapping trials on real-world data in two different environments. In (a), the robot successfully built a map which accurately represents the underlying environment. In (b), despite a small alcove with challenging lighting for which no training data was collected, the learned sensor was able to accurately build much of this atrium environment.
    %However, in (b), due to harsh lighting conditions not seen during training, the learned sensor was unable to accurately build one of the branching hallways in the building. Given sufficient training however, we would expect to overcome this failure.
    }\label{fig:results:real_world}
\end{figure}

\section{Mapping with Real Monocular Camera Data}
\label{sec:robot_experiments}

\begin{comment}
- How do we make training data
- What do we do testing on
- What do the results show
    - robust to types of failure modes in other monocular systems on real data (figure comparing to VSLAM)
    - some element of repeatability + shown in other environments
\end{comment}

To demonstrate our learned sensor and mapping procedure on real data, a teleoperated robot equipped with a Ricoh Theta S\cite{ThetaCamera} panoramic camera and Hokuyo \textsc{lidar}\cite{Hokuyo} (used to generate training data) was driven in three distinct indoor environments on the MIT campus.
%: Building 6C Atrium, Building 36-2, and Building E52-2. 
We drove the robot through our target environments multiple times (fewer than 10 traversals for each environment; on the order of 10k images) and trained a sensor for each environment. We generated ground truth occupancy grids with Cartographer\cite{cartographer} and hand-built the polygonal map of each environment with clutter removed to pair with localized images for training.
% , we generated ``ground truth'' 2D occupancy maps from which we built hand-labeled polygons of each environment, which we used to mark vertices and walls in localized panoramic images taken by the camera. 
The training and testing sets were mutually exclusive, though the generalization performance of our planner to unseen environments is the subject of future work.

% To collect training and testing data-sets for our learned sensor, the robot was driven multiple times through each of these environments. Google Cartographer\cite{cartographer} was subsequently used to construct ``ground truth'' 2D occupancy maps of each environment from Lidar measurements and generate a localized trajectory for the robot within the maps. For each map, a hand-labeled polygonal map with vertex and edge information was created to define points of high-level structure, which ignored small obstacles and clutter that appears in the occupancy grid generated from the laser data.  For each panoramic image, the visible vertices and edges in the polygonal map were extracted by ray-casting from the localized pose of the robot yielded by Cartographer. 

%We show results in Figs.~\ref{fig:results:real-world-front-page} and \ref{fig:results:real_world}:
\paragraph{Bld. 36-2, Fig.~\ref{fig:results:real_world}\texttt{(a)}} Our first environment was a simply-connected hallway intersection. Despite a high number of false-positive detections (red circles), we correctly built the underlying map, capturing the high-level structure and building a single contiguous boundary.

\paragraph{Bld. 6C Atrium, Fig.~\ref{fig:results:real_world}\texttt{(b)}} This environment consisted of one large loop and a small annex with large windows and glass walls. Although we were able to correctly map the central loop, we failed to accurately represent the annex, where light glare saturated the images. Notably, during the trial, the robot at one point had an accurate representation of the annex (highlighted in Fig.~\ref{fig:results:real_world}\texttt{(b)}), yet the challenges in detection overwhelmed and resulted in many false positive detections (red circles) in this region.
% Many false positive detections (red circles) in the area of the annex is indicative of the difficulty the detector had under these conditions. 

\paragraph{Bld. E52-2, Fig.~\ref{fig:results:real-world-front-page}} This environment consisted of two loops and several branching hallways populated with clutter (benches, chairs) and challenging surfaces (glass and textureless walls). In these trials we show that we are able to consistently build a map which, qualitatively, well represents the underlying structure of the environment. Quantitatively, 10 of 11 trials built the two loops in the environment completely and accurately. In the unsuccessful trial, there was a failure of data association in the rightmost loop; where there should be 1 vertex, there were 3 rejected vertices (red circles). As compared to a state-of-the-art monocular \textsc{slam} package \cite{openvslam2019} for panoramic input, highlighted in Fig.~\ref{fig:results:real-world-front-page}, our learned sensor performed robustly in predominantly textureless regions.

\section{Conclusion}
In this work, we present a sparse map that enables planning with topological constraints using monocular vision. We introduce a learned sensor to detect vertices from panoramic images and estimate the presence of large obstacles that connect them in order to extract a polygonal representation of the environment that includes only high-level structure. Our map also includes an estimate of known free space, which we use to define topological constraints---e.g., homology and frontier constraints---for planning through unknown environments. Upon completion, our procedure yields a polygonal map which is sufficient for optimal navigation via computation of a visibility graph. Our results motivate future work that generalizes to novel environments, extends to actions beyond the domain of navigation, and incorporates localization within our representation.

We demonstrate the utility and robustness of our representation in both \emph{multi-agent search} and \emph{visual learned subgoal planning}. In simulated trials, we show that our representation can be used to efficiently search unknown space. Moreover, we demonstrate that our map is sufficiently stable to enable visual learned subgoal planning. Through trials in several representative indoor environments, we further show that our map can be generated from real, noisy visual data under conditions in which other methods relying on sparse visual features have been shown to fail. 

%===============================================================================

% The maximum paper length is 8 pages excluding references and acknowledgements, and 10 pages including references and acknowledgements

% \clearpage
% The acknowledgments are automatically included only in the final version of the paper.

%===============================================================================

\bibliographystyle{IEEEtran}
\bibliography{root}  % .bib

%%%%%%%%%%%%%%%%%%%%%%%%%%%%%%%%%%%%%%%%%%%%%%%%%%%%%%%%%%%%%%%%%%%%%%%%%%%%%%%%%%%%%%
%%%%%%%%%%%%%%%%%%%%%%%%%%%%%%%%%%%%%%%%%%%%%%%%%%%%%%%%%%%%%%%%%%%%%%%%%%%%%%%%%%%%%%
% START OF PREVIOUS DRAFT
%%%%%%%%%%%%%%%%%%%%%%%%%%%%%%%%%%%%%%%%%%%%%%%%%%%%%%%%%%%%%%%%%%%%%%%%%%%%%%%%%%%%%%
%%%%%%%%%%%%%%%%%%%%%%%%%%%%%%%%%%%%%%%%%%%%%%%%%%%%%%%%%%%%%%%%%%%%%%%%%%%%%%%%%%%%%%

\end{document}